\documentclass[conference,10pt]{IEEEtran}
\IEEEoverridecommandlockouts
\usepackage{cite}
\usepackage{amsmath,amssymb,amsfonts}
\usepackage{algorithmic}
\usepackage{graphicx}
\usepackage{textcomp}
\usepackage{xcolor}
\def\BibTeX{{\rm B\kern-.05em{\sc i\kern-.025em b}\kern-.08em
    T\kern-.1667em\lower.7ex\hbox{E}\kern-.125emX}}
\usepackage[linesnumbered,lined,boxed,commentsnumbered,ruled,longend]{algorithm2e} 
\usepackage{bm} 
\usepackage{bbold} 
\graphicspath{{./Figs/}}
\usepackage[caption=false, font=footnotesize]{subfig} 
\usepackage{multirow}

\SetKwInput{KwData}{Input}
\SetKwInput{KwResult}{Output}
\SetKwProg{Init}{Initialize}{}{}
\SetKwFor{ParFor}{for}{do in parallel}{end for}
\SetKwRepeat{Do}{do}{while}

\newcommand{\cache}[1]{} 

\begin{document}

\title{Battery-aware Cyclic Scheduling in Energy-harvesting Federated Learning\\
\thanks{This work was supported by the Horizon Europe/JU SNS project ROBUST-6G (Grant Agreement no. 101139068). The work of N. Pappas has been supported in part by the Swedish Research Council (VR), ELLIIT, and the EU (ETHER, 101096526, ELIXIRION, 101120135, and SOVEREIGN, 101131481).}
}

\author{\IEEEauthorblockN{Eunjeong Jeong and Nikolaos Pappas}
\IEEEauthorblockA{\textit{Department of Computer and Information Science}, 
\textit{Linköping University, Sweden}\\
\texttt{\{eunjeong.jeong, nikolaos.pappas\}@liu.se}}
}

\maketitle

\begin{abstract}
    Federated Learning (FL) has emerged as a promising framework for distributed learning, but its growing complexity has led to significant energy consumption, particularly from computations on the client side. This challenge is especially critical in energy-harvesting FL (EHFL) systems, where device availability fluctuates due to limited and time-varying energy resources. We propose FedBacys, a battery-aware FL framework that introduces cyclic client participation based on users’ battery levels to cope with these issues. FedBacys enables clients to save energy and strategically perform local training just before their designated transmission time by clustering clients and scheduling their involvement sequentially. This design minimizes redundant computation, reduces system-wide energy usage, and improves learning stability. Our experiments demonstrate that FedBacys outperforms existing approaches in terms of energy efficiency and performance consistency, exhibiting robustness even under non-i.i.d. training data distributions and with very infrequent battery charging. This work presents the first comprehensive evaluation of cyclic client participation in EHFL, incorporating both communication and computation costs into a unified, resource-aware scheduling strategy.
\end{abstract}


\section{Introduction}\label{sect:intro}

Federated Learning (FL) \cite{konevcny16:FL} has gained widespread attention for enabling efficient and privacy-preserving distributed learning, with its applications rapidly expanding to tackle increasingly complex tasks \cite{li20:magazine}. However, this growth has brought a global concern for excessive energy consumption in FL. In particular, client-side computations constitute the most significant portion of energy usage across all FL stages \cite{yousefpour23:greenFL, thakur25:greenFL}. Optimizing the user schedule is essential to balance FL performance and energy consumption.

Effective client selection is also critical from a wireless communication perspective, especially in large-scale networks, as it helps to alleviate congestion, reduce collisions, and prevent interference. While many existing client selection algorithms have pointed out unreliable channel conditions that primarily limit the number of available clients for model aggregation, they often overlook the battery levels of each client~\cite{hamdi22, xu21, albelaihi22:greenFL}. Scheduling policies often consider users' channel conditions. However, if FL schemes assume that all users have abundant energy resources and are available to join whenever they want, their algorithms cause unstable performances when the battery levels of each client are time-variant. Thus, energy harvesting needs to be considered while designing distributed learning algorithms.

Considering the limitations of traditional FL in energy-constrained environments, Energy-harvesting Federated Learning (EHFL) has emerged as a promising framework. EHFL aims to extend the operational lifespan of its clients and enhance learning performance by harnessing ambient energy to continuously power edge devices. Primarily, EHFL research has focused on maximizing the number of participating clients in each round to minimize training loss \cite{shen22}. The scheduling of devices in EHFL is often jointly optimized with resource allocation strategies \cite{zeng23}, power control \cite{zhang24}, or information privacy \cite{pan22}. Nonetheless, most existing works still adhere to the conventional FL architecture where each client directly communicates with the central server \cite{guler21, aygun22, poposka24}. This dependence on direct client-server transmission implies potential improvements in the communication architecture, particularly considering that users are inherently unreliable and their availability fluctuates due to their time-varying energy reserves. 

In this regard, to address the challenges posed by energy constraints and the limitations of existing EHFL approaches, we propose FedBacys, a battery-aware FL scheme with cyclic client participation. FedBacys enhances energy efficiency by leveraging user clustering based on their remaining battery levels and sequentially aggregating intra-group models within a defined cycle. In this scheme, users are assigned to several groups with sequentially scheduled participation times, prompting them to reserve sufficient energy until their designated turn for uploading updates approaches. This timely task scheduling, where users strategically delay local training until shortly before their scheduled update transmission, effectively minimizes redundant client-side computations and contributes to more stable performance by regulating the number of active participants in each communication round.

Prior works have indicated that sequential model aggregations over user clusters can achieve faster asymptotic convergence rates compared to standard FedAvg under certain conditions \cite{cho23}, and cyclic participation has been suggested for resource-heterogeneous networks. Additionally, grouping clients for in-cluster model aggregation reduces the total communication cost by limiting the number of users directly uploading to the server \cite{chen23:FedSeq, lee23}. However, these existing works primarily focus on communication efficiency and often overlook the significant energy consumed by client-side computations, particularly the cost of local model training. 

To the best of our knowledge, this paper presents the first comprehensive investigation into the robustness and effectiveness of cyclic client participation within an EHFL framework, explicitly considering both communication and computation costs through a resource-aware criterion for clustering clients.



\section{System Model}\label{sect:system_model}
\begin{figure}[t]
    \centering
    \includegraphics[width=\columnwidth]{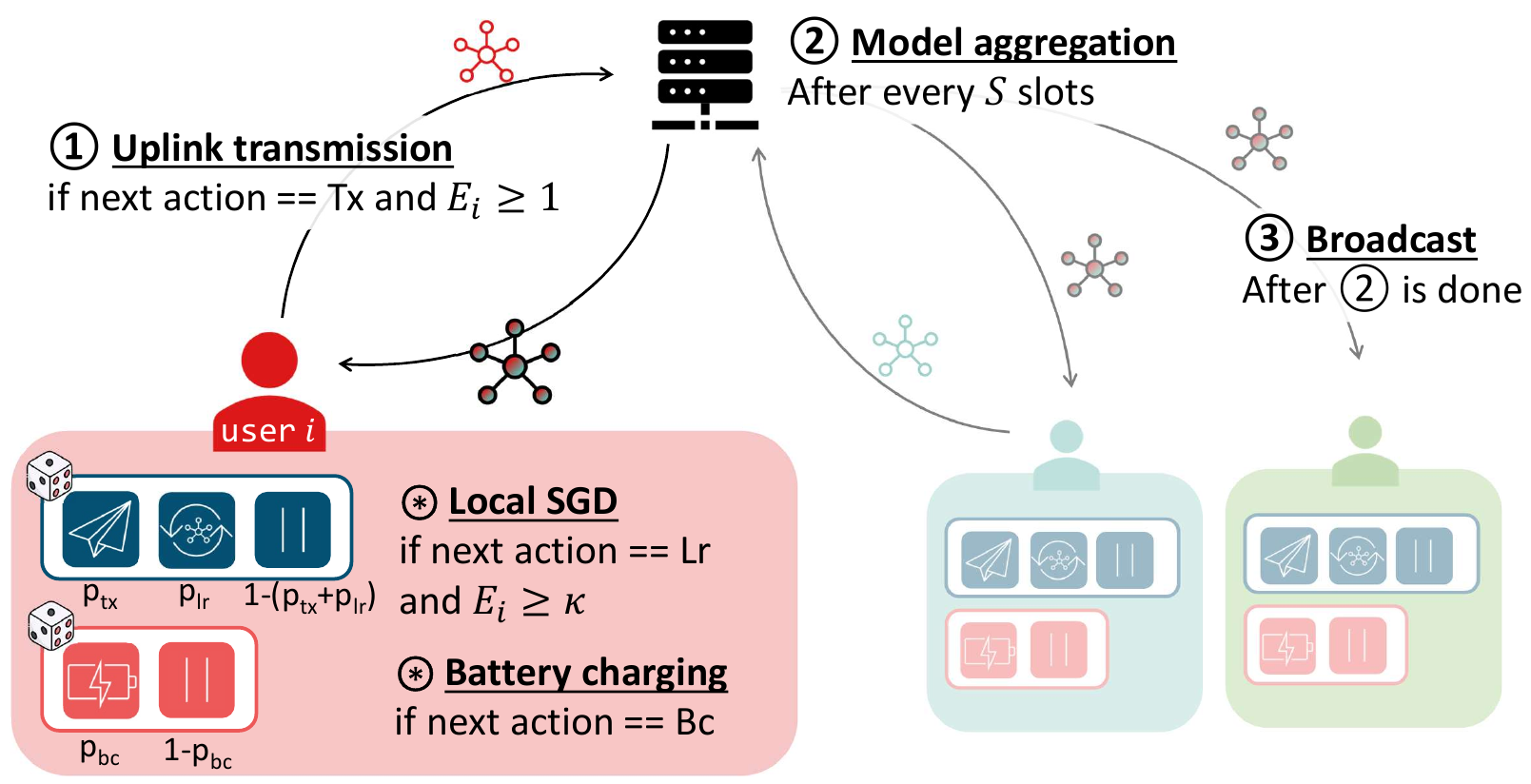}
    \caption{A schematic view of energy-harvesting FL based on probabilistic decision rules.}
    \label{fig:prob_ehfl}
\end{figure}

We consider the following optimization task over $N$ clients whose goal is to minimize
\begin{align}
    f(\mathbf{x}) := \frac{1}{N}\sum_{i\in\mathcal{U}} f_i(\mathbf{x})
    \label{fn:obj}
\end{align}
where $\mathbf{x}\in \mathbb{R}^d$ is an $d$-dimensional model parameter, $f_i$ is a local loss function, and $\mathcal{U}=\{1,\cdots,N\}$ is a set of network users. Each client possesses a local training dataset that is not shared with other clients. 

Energy-harvesting Federated Learning (EHFL) follows the standard FL workflow composed of server-side computation, downlink communication, client-side computation, and uplink communication. Separated from the communication and computation stages, each client in EHFL independently harvests and stores energy in an energy queue, which is modeled similarly to \cite{shen22}. We assume normalized energy units, such that activity taking one time slot requires one battery unit to be spent. The system operates in discrete time slots, with each unit of time corresponding to a single scheduling interval.

In each time slot, a user can decide to perform one of three actions, provided that the user is not currently engaged in a previous task: (1) transmitting, (2) training its local model, or (3) staying idle, as illustrated in Fig.~\ref{fig:prob_ehfl}. The action of transmission or being idle occupy one time slot. 
In contrast, a client that initiates local model training remains busy for the subsequent $\kappa$ time slots and cannot make further decisions until this task is complete.
More specifically, let $\mathcal{T}_s^{(i)}\in\{0,1\}$ be a binary indicator such that $\mathcal{T}_s^{(i)}=1$ if user $i$ transmits at slot $s$, and $\mathcal{T}_s^{(i)}=0$ otherwise. Similarly, let $\mathcal{L}_s^{(i)}\in\{0,1\}$ denote whether user $i$ performs local model training at slot $s$. Clients cannot simultaneously transmit a message and train their local model, thus we have $\mathcal{T}_s^{(i)}+\mathcal{L}_s^{(i)} \leq 1$.
Meanwhile, the battery charging event of each user $i$ at each slot $s$, denoted as $\mathcal{C}_s^{(i)}$, follows a Bernoulli distribution with probability \(p_{bc}=\text{Pr}[\mathcal{C}_s^{(i)}=1]\), where $\mathcal{C}_s^{(i)}=1$ if user $i$ charges its battery at slot $s$ and $\mathcal{C}_s^{(i)}=0$ otherwise. 

Harvesting and energy-consuming actions affect the dynamics of the battery level. Transmitting requires one battery unit and occupies one time slot. When user $i$ decides to transmit a message, its battery level evolves as
\begin{equation}
    E_{s+1}^{(i)} = \max(E_s^{(i)}-1,0)+\mathbb{1}_{\{\mathcal{C}_s^{(i)}\}},
\end{equation}
where $\mathbb{1}$ is the indicator function.
A single occurrence of local model training consumes $\kappa$ battery units and spans $\kappa$ consecutive time slots. Thus, the battery level evolution in this case, after $\kappa$ slots is given by
\begin{equation}
    E_{s+\kappa}^{(i)} = \max(E_s^{(i)}-\kappa,0)+\sum_{s'=s}^{s+\kappa-1}\mathbb{1}_{\{\mathcal{C}_{s'}^{(i)}\}}.
\end{equation}
The battery level of the user staying idle at slot $s$ evolves as
\begin{equation}
    E_{s+1}^{(i)} = E_s^{(i)}+\mathbb{1}_{\{\mathcal{C}_s^{(i)}\}}.
\end{equation}

The users cannot perform energy-consuming actions, such as local model training or transmitting, when their battery level is $0$. A client can only undertake an action if it has sufficient energy to complete it; if an action consumes more energy than currently available, it is declined.  

\cache{
\begin{equation}
    E_{s+1}^{(i)}=\max\Big(E_s^{(i)}-\big(\mathbb{1}_{\{\mathcal{T}_s^{(i)}\}} +\sum_{s'=s-\kappa}^s\mathbb{1}_{\{\mathcal{L}_{s'}^{(i)}\}}\big),\ 0 \Big)+\mathbb{1}_{\{\mathcal{C}_s^{(i)}\}}
\end{equation}
}

\section{Energy-harvesting Cyclic FL}\label{sect:fedbacys}

\begin{figure}[t]
    \centering
    \includegraphics[width=\columnwidth]{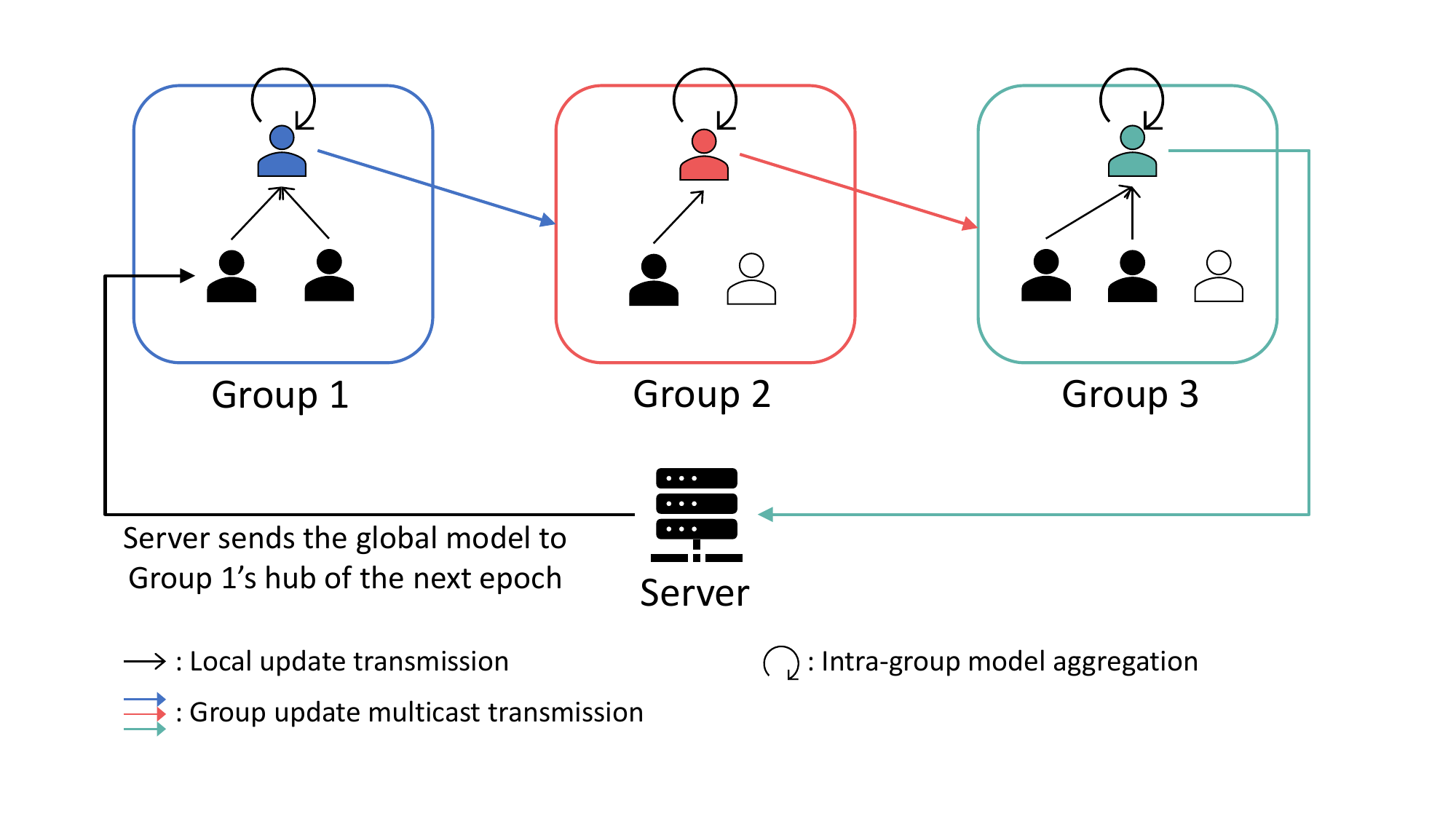}
    \caption{A simplified example of a FedBacys network with $N=10$ users assigned into $G=3$ groups. Within each group, (1) each user transmits local updates to the group's hub if they satisfy the conditions to send local updates; (2) the hub aggregates the received updates; (3) the hub sends the updated intra-group model to the next group by multicast. The final group's hub sends the aggregated model to the server, which is then sent to Group 1 of the new epoch. The hub users can be changed at the beginning of every epoch.}
    \label{fig:FedBacys}
\end{figure}

In FedBacys, two time units are defined: \emph{slot} and \emph{epoch}. A slot represents a time step where users determine and perform an action. An epoch, on the other hand, corresponds to a complete communication round where the server aggregates local model updates to refine the global model. An epoch consists of $S$ slots. For example, when $S=30$ and the clients are divided into $G=3$ groups, the aggregation process within each group takes $R=\lfloor \frac{S}{G} \rfloor = 10$ slots sequentially, where $R$ denotes the duration of one group-round. 

During each slot, every user has a probability $p_{bc}$ of harvesting one battery unit. Before the collaborative learning process, the server assigns users to one of the $G$ clusters (groups). At slot $s$, user $i$ determines when to start training its local model according to the following three conditions:
\begin{enumerate}
    \item Sufficient battery: the user has enough battery to train locally, i.e., $E_s^{(i)} \geq \kappa$
    \item No pending update: the user does not yet have a local update $\Delta^{(i)}$ to upload.
    \item Upload deadline observance: the user can finish local training within its group's active period, i.e.,
    \begin{equation*}
        g_iR\leq s+\kappa\ (\text{mod } S) <(g_i+1)R-1.
    \end{equation*}
    where $g_i$ is the group assigned to user $i$.
\end{enumerate}

The third condition implies that users begin local model training only when they can barely meet the deadline for uploading their update. User $i$ assigned to group $g_i$ can transmit its local update to the group hub during the time window between two slots, whose remainders after division by $S$ are $g_iR$ and $(g_i+1)R-1$, respectively. If this user knows the number of slots required to complete the local training, denoted by $\kappa$, it is sufficient to begin training $\kappa$ slots ahead of the aforementioned time window. In other words, users do not need to launch local training when the first two conditions are met. This ``cramming'' behavior helps users save time by reducing computation costs and accelerating global consensus. First, participants train local models less frequently since those who can afford battery levels may stay idle until the last slot to begin training and meet the upload deadline. Second, users may have access to a fresher global model until they begin local training, enabling them to contribute more to the global update on the upcoming epochs.

Clients are evenly distributed across the remaining clusters, ensuring each group has an almost equal number of users. A temporary hub of each group for intra-group model aggregation is randomly chosen. At the final slot of each group round, the hub collects local updates from the group members and delivers the intermediate global model to the next group by multicasting. (Lines 12-17 of Alg.~\ref{alg:FedBacys}) The hub of the last group transmits the global model to the server. At the beginning of the next epoch, the server reassigns the group hubs and sends its latest global model to the first group. 



\begin{algorithm}[tb]
    \SetAlgoLined \SetAlgoNoEnd
    \caption{\label{alg:FedBacys} Energy Harvesting FedAvg with Cyclic Client Participation}
    \DontPrintSemicolon
    \KwData{$T,\ \mathcal{U},\ G,\ E_0^{(i)}=0\ \forall i\in\mathcal{U}, \ p_{bc}$,
    local training model $\mathbf{x}_0^{(i)}=\mathbf{y}_0^{(i)}=\mathbb{0}$ $\forall i\in\mathcal{U}$, learning rate $\eta$}
    \KwResult{$\{\mathbf{x}_t : \forall t\}$}
    \SetKwFunction{LocalTrain}{LocalTrain}
    \SetKwFunction{Grouping}{Grouping}
    \SetKwProg{Fn}{Function}{:}{end}

    \For{$s=0,\cdots,ST-1$}{
        \If{$s=0$}{
        $\mathcal{N}_1,\cdots,\mathcal{N}_G \leftarrow$ Randomly sliced $\mathcal{U}$ \; 
            Pick $i^*_1,\cdots,i^*_G$ among $\mathcal{N}_1,\cdots,\mathcal{N}_G$ \tcp*{Randomly choosing group hub}
        }
        $t\leftarrow \lfloor s/S\rfloor$ \tcp*{index of current epoch}
        $g\leftarrow \lfloor (s\mod S)/R\rfloor$ \tcp*{index of current group}
        \ParFor{each $i\in \mathcal{U}$}{
            \If{$\mathcal{C}_s^{(i)}=1$}{
                $E_s^{(i)}\leftarrow E_s^{(i)}+1$ \tcp*{Charge battery}
            }
            \If{$i$ satisfies learning conditions}{
                \LocalTrain{$\mathbf{x}^{(i)}_t,\eta,B$}
            }
        }
        \If{$s\equiv R-1\ (\text{mod } R)$}{
            \For{$i\in\mathcal{N}_g$}{
                \If{$E_s^{(i)}\geq1$ and $i$ has $\Delta_t^{(i)}$ that has not been uploaded yet}{
                    Send $\Delta_t^{(i)}$ to $i^*_g$ \tcp*{intra-group model aggregation}
                }}
            $\mathbf{x}_{t}^{(i^*)} \leftarrow \mathbf{x}_{t}^{(i^*)} + \sum_{i\in\mathcal{N}_g}\Delta_t^{(i)}$ \;
            $\mathbf{x}_{t,g} \leftarrow \mathbf{x}_{t}^{(i^*)}$ \;
            \For{$j\in\mathcal{N}_{g+1}$}{
                \If{$g<G$}{
                    $i^*_g$ sends $\mathbf{x}_{t,g}$ to $j\in\mathcal{N}_{g+1}$ by multicast\;
                    $\mathbf{x}_t^{(j)}\leftarrow \mathbf{x}_{t,g}$ \;
                }
                \Else{
                    $i^*_g$ sends $\mathbf{x}_{t,g}$ to Server \;
                    $\mathbf{x}_{t+1, 0}\leftarrow \mathbf{x}_{t, G}$\; 
                }
            }
        }
    }
    \Return $\mathbf{x}_T$\;
    
    \BlankLine
    \Fn{\LocalTrain{$\mathbf{x}^{(i)}_t,\eta,B$}}{
        \For{each batch $b=0,\cdots,B-1$}{
                $\mathbf{y}_{t,b+1}^{(i)} \leftarrow \mathbf{y}_{t,b}^{(i)} - \eta \nabla f_i(\mathbf{y}_{t,b}^{(i)})$ \; 
                $\Delta_t^{(i)}\leftarrow\mathbf{x}_t^{(i)}- \mathbf{y}_{t,B}^{(i)}$ \; 
            }
    }\Return $\Delta_t^{(i)}$\; 
\end{algorithm}



\section{Experimental Results}\label{sect:exp}

\begin{figure}[t]
    \centering
    \includegraphics[width=0.93\columnwidth]{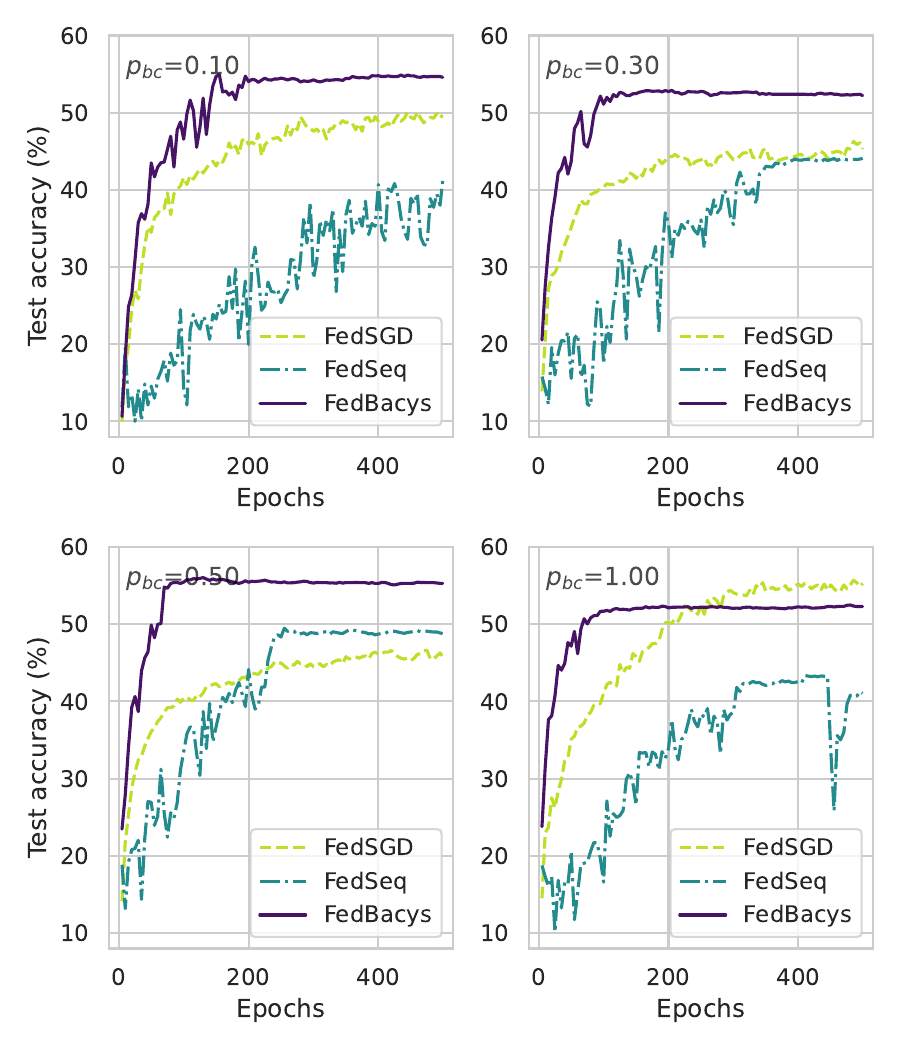}
  \caption{Test accuracy w.r.t. different algorithms ($G=10$ groups)}
  \label{plot:main} 
\end{figure}

To evaluate FedBacys, we conduct image classification experiments using the CIFAR-10 dataset. The training model employed is a neural network consisting of two convolutional layers, one max-pooling layer, and three fully connected layers. Each user has 50 training samples, totaling 100 users ($N=100$). Local model training utilizes stochastic gradient descent (SGD) with a learning rate of  $\eta=0.05$, and the cross-entropy function is used to calculate training losses. The training runs for 500 epochs ($T=500$), each comprising $S=30$ slots.
We explore the impact of battery charging probability by setting $p_{bc} = \{ 0.1, 0.3, 0.5, 1.0 \}$. The computational cost for a single local training session is fixed at $\kappa=20$ battery units. The maximum battery capacity is set to $E_{\text{max}} = \kappa + 5$, ensuring that successive local training without transmission is avoided. Each uplink transmission consumes one battery unit, while local model training requires $\kappa = 20$ battery units. The initial battery state for all users is set to $E_0^{(i)} = 0$.

We compare our proposed scheme with FedAvg~\cite{mcmahan17:fedavg} and FedSeq~\cite{chen23:FedSeq}. Since the comparisons do not have system models for energy-harvesting settings, we implement the experiments under the same battery charging conditions in which users harvest energy to perform transmission and local model training. Instead of choosing a subset for participation in each epoch as in traditional FedAvg, all $N$ users are considered potential participants in every epoch. Still, some might fail to join due to a lack of battery or unprepared local updates at the aggregation moment. 


Fig.~\ref{plot:main} presents the average test accuracy across epochs for various algorithms. FedAvg demonstrates consistent and stable performance regardless of $p_{bc}$ and $G$. In contrast, FedSeq and FedBacys tend to fluctuate before reaching their saturation points because of the simultaneous presence of users who have completed global model updates and those who are processing or queued for updates within the same epoch. Across all schemes, higher battery charging probabilities lead to a higher convergence rate. Notably, FedBacys reaches both faster convergence and higher final test accuracies for $p_{bc}$ values of 
0.1, 0.3, and 0.5. Yet, FedAvg yields the highest final average test accuracy when $p_{bc}=1.0$. 

\begin{figure}[t]
    \centering
    \includegraphics[width=0.93\columnwidth]{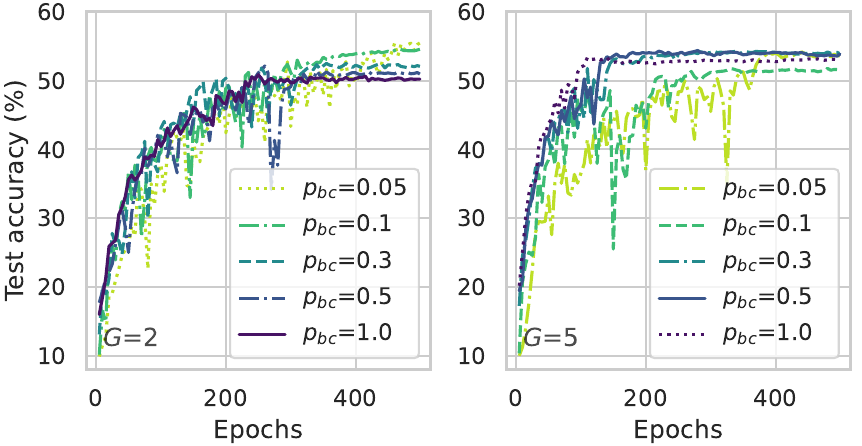}
    \caption{Average test accuracy of FedBacys w.r.t. different battery charging probabilities. ($G=2,\ 5$)}
    \label{plot:battery}
\end{figure}

The impact of battery charging probability on the performance of FedBacys networks is illustrated in Fig.~\ref{plot:battery}. When the $p_{bc}$ is low, increasing the number of clustering groups ($G$) results in more severe oscillations in the accuracy curves. However, when the energy is reliably supplied due to high battery charging probabilities, the final average test accuracy, convergence speed, and total energy consumption (as detailed in Table~\ref{table:energy-consumption}) remain consistently high regardless of the number of groups.


We compare the total energy consumption across the entire network for different schemes with three clustering setups with $G=\{2,5,10\}$. Note that the number of groups is not considered in FedAvg since the algorithm does not have clustering policies for participant selection. Table~\ref{table:energy-consumption} reveals that FedBacys consumes the least energy across all battery charging probabilities. Combined with the accuracy results presented in Fig.~\ref{plot:main}, FedBacys provides the highest performance per battery usage no matter how frequently the battery level is increased in each client.
The table also demonstrates that the gap in computation costs between FedBacys and the other schemes is small when $p_{bc}$ is low. However, FedBacys achieves energy savings of $26-30$\% of energy when $p_{bc}$ is high. This disparity in energy efficiency across different $p_{bc}$ occurs because of the inherent limitations of battery scarcity, which is an unavoidable condition to practice actions. The minimum number of local training attempts required to obtain global consensus is constrained when energy is severely scarce, regardless of the chosen scheme.
As a result, under limited battery availability, all schemes exhibit similar energy consumption due to the fundamental constraint on training frequency. However, as energy availability increases, FedBacys leverages its battery-aware scheduling to significantly reduce client-side computation costs without sacrificing accuracy, demonstrating its effectiveness in optimizing learning performance and energy efficiency across varying battery conditions.

\begin{table}[t]
\centering
\caption{\label{table:energy-consumption} Total energy consumption of different FL schemes under various $p_{bc}$ and $G$.}
\begin{tabular}{|l|c|cccc|}
\hline
                        & \multirow{2}{*}{$G$}   & \multicolumn{4}{c|}{$p_{bc}$}      \\ \cline{3-6} 
                        &      & 0.1             & 0.3             & 0.5    & 1.0     \\ \hline
FedAvg                  & -    & 149232          & 448931          & 747768 & 1498100 \\ \hline
\multirow{3}{*}{FedSeq} & 2    & 149056          & 448846          & 749233 & 1488369 \\ \cline{2-6} 
                        & 5    & 149524          & 448617          & 748317 & 1498100 \\ \cline{2-6} 
                        & 10   & 149132          & 449010          & 748133 & 1433830 \\ \hline
\multirow{3}{*}{FedBacys}   & 2    & 148665          & 444857          & 728421 & 1048750 \\ \cline{2-6} 
                        & 5    & \textbf{148541} & 434010          & 649471 & 1048280 \\ \cline{2-6} 
                        & 10   & 148686          & \textbf{425484} & \textbf{608659} & \textbf{1047970} \\ \hline
\end{tabular}
\end{table}

\begin{figure}[t]
    \centering
    \includegraphics[width=0.82\columnwidth]{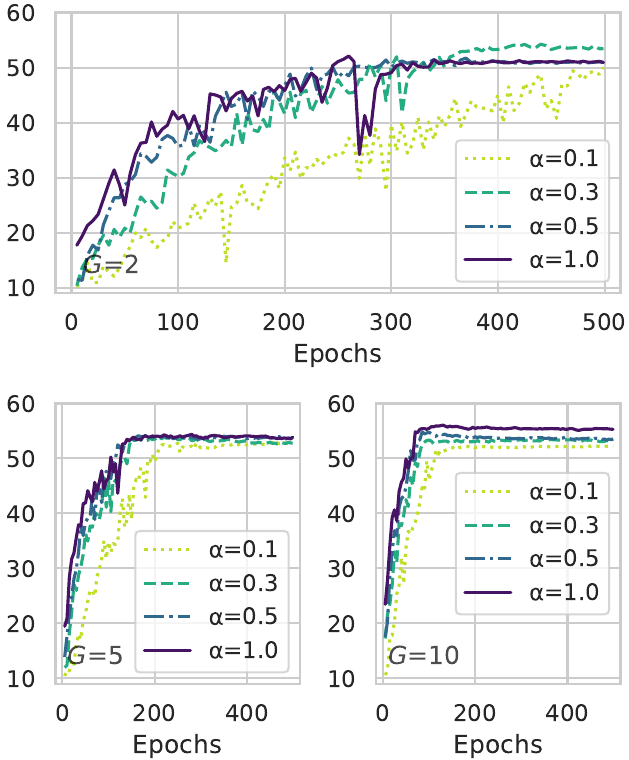}
    \caption{Average test accuracy of FedBacys on non-i.i.d. training datasets with different Dirichlet coefficients applied. $p_{bc}$ is fixed to 0.5 for all trials.}
    \label{plot:non-iid}
\end{figure}
While most of our experiments were conducted under i.i.d. data settings, we also evaluated FedBacys with non-i.i.d. training datasets, as illustrated in Fig.~\ref{plot:non-iid}. As the Dirichlet coefficient decreases, which indicates higher variability in local training data distributions, the convergence rate decreases, and the fluctuations in the accuracy curves before saturation become more prominent.
Notably, however, FedBacys delivered more robust performances against non-i.i.d. data distributions when users were assigned to a larger number of groups ($G=10$).

\section{Conclusions and Future Directions}\label{sect:conc}


This paper presented \textit{FedBacys}, a battery-aware FL scheme that enhances energy efficiency and performance stability in energy-harvesting scenarios through cyclic client participation. The proposed method promotes a timely local training strategy by minimizing redundant computations via user clustering based on remaining battery levels and sequential intra-group model aggregation. Simulation results demonstrate that FedBacys significantly improves the stability of FL performance and reduces energy consumption at both client and system levels, particularly under fluctuating energy availability. By explicitly accounting for communication and computation costs through a resource-aware clustering criterion, FedBacys addresses a critical gap in existing EHFL research, which often neglects client-side training energy expense.

Future work may explore the integration of semantics-aware client participation strategies~\cite{delfani24, hu24:vAoI-FL} to further optimize resource utilization and mitigate superfluous computations and communications in EHFL systems. Moreover, investigating the interdependencies among key hyperparameters, such as maximum battery capacity, client-side computation cost, and battery charging probabilities, offers a promising direction for extending and fine-tuning the proposed framework. A deeper understanding of these relationships could enable developing more adaptive and efficient EHFL systems across diverse real-world applications.

\bibliographystyle{ieeetr}  
\bibliography{ref}

\end{document}